\pdfoutput=1

\documentclass[11pt]{article}

\usepackage[]{acl}

\usepackage{times}
\usepackage{latexsym}

\usepackage[T1]{fontenc}

\usepackage[utf8]{inputenc}

\usepackage{microtype}

\usepackage{inconsolata}

\usepackage{amsmath} 
\usepackage{amsfonts}
\usepackage{multirow}
\usepackage{graphicx}
\usepackage{hhline}
\usepackage{booktabs}

\DeclareMathOperator*{\argmax}{arg\,max}

\title{Direct Preference Optimization for Neural Machine Translation with Minimum Bayes Risk Decoding}

\author{Guangyu Yang, Jinghong Chen, Weizhe Lin, Bill Byrne \\
    Department of Engineering \\ University of Cambridge \\
    \texttt{\{gy266, jc2124, wl356, wjb31\}@cam.ac.uk}
    }

\begin{document}
\maketitle
\begin{abstract}
Minimum Bayes Risk (MBR) decoding can significantly improve translation performance of Multilingual Large Language Models (MLLMs). However, MBR decoding is computationally expensive.  We show how the recently developed Reinforcement Learning technique, Direct Preference Optimization (DPO), can fine-tune MLLMs to get the gains of MBR without any additional computation in inference.  Our method uses only a small monolingual fine-tuning set and yields significantly improved performance on multiple NMT test sets compared to  MLLMs without DPO. 
\end{abstract}

\section{Introduction}

MBR decoding \cite{kumar-byrne-2004-minimum, eikema-aziz-2022-sampling, suzgun-etal-2023-follow} is a two-pass procedure that generates multiple translation hypotheses and selects a hypothesis based on Bayesian risk. Recent work \cite{garcia2023unreasonable, suzgun-etal-2023-follow, yang-2023-thesis} has shown that MBR decoding can significantly boost the translation performance of MLLMs \cite{lin-etal-2022-shot, muennighoff-etal-2023-crosslingual, zeng2023glm130b}, outperforming greedy decoding and beam search. However, MBR decoding is expensive, both in computation and in latency.

Our goal is to fine-tune a base MLLM so that it has the same single-pass decoding performance as MBR decoding. We propose
a novel self-supervised fine-tuning method based on DPO~\cite{DBLP:conf/nips/RafailovSMMEF23}. Our method uses MBR decoding on an MLLM to produce a preference dataset consisting of pairs of ranked translations. The DPO algorithm is used to fine-tune the MLLM to prefer the higher-ranked translations over lower-ranked ones. MLLMs optimized for MBR preference achieve significantly better translation performance when decoded with beam search, achieving translation quality on par with MBR decoding of the original model.

\section{MBR and DPO}



We follow the expectation-by-sampling approach to MBR  \cite{eikema-aziz-2022-sampling}.  Given a set of sampled translations $H({\bf x}) = \{ {\bf y}' \sim P(\cdot | {\bf x})\}$ and a  loss (or utility) function $L(\cdot, \cdot)$, the score (negative Bayes risk) of each translation is found as 
\begin{equation}
S({\bf y}) =  - \frac{1}{|H(x)|} \sum_{{\bf y} '\in H({\bf x})} L({\bf y}', {\bf y})
\label{eq:riskscore}
\end{equation}
and the MBR hypothesis is then computed as 
\begin{equation}
\mathbf{y}^* = \argmax_{\mathbf{y}\in H({\bf x})}S(\mathbf{y}) 
\end{equation}
This is simple but  expensive.   Our goal is to train a model that produces translations with scores consistent with MBR, but without  multi-step decoding.

%
%
%



\subsection{DPO Fine-Tuning Objective} \label{sec:dpodescrip}
DPO \cite{DBLP:conf/nips/RafailovSMMEF23} reformulates the usual approach to Reinforcement Learning from Human Feedback (RLHF) 
so as to avoid a distinct reward modelling step.   The typical RLHF criteria is 
\begin{align}
\max_{\pi_{\theta}}~ &\mathbb{E}_{\mathbf{x} \sim D, \mathbf{y} \sim \pi_{\theta}(\mathbf{y}|\mathbf{x})} \left[ r_{\phi}(\mathbf{x}, \mathbf{y}) \right] \\
&- \beta \mathbb{D}_{KL} \left[ \pi_{\theta}(\mathbf{y} | \mathbf{x}) \parallel \pi_{\text{ref}}(\mathbf{y} | \mathbf{x}) \right] \nonumber
\end{align}
where $r_{\phi}$ is a reward model trained from human feedback, $\pi_{\theta}$ is the model being trained, and $\pi_{\text{ref}}$ is the reference model.   DPO effectively replaces the reward model with a preference distribution based on $\pi_\theta$, the model being trained;  DPO also retains the KL regularization term with weighting $\beta$.

The preference dataset $D$ for DPO consists of  triplets $(\mathbf{x}, \mathbf{y}_w, \mathbf{y}_l)$ where $\mathbf{x}$ is the input prompt, $\mathbf{y}_w$ is the winnng (prefered) response, and $\mathbf{y}_l$ is the losing (disprefered) response. DPO uses the language model likelihood to approximate the reward as $\beta\text{log}\frac{\pi_{\theta}(\mathbf{y}|\mathbf{x})}{\pi_{\text{ref}}(\mathbf{y}|\mathbf{x})}$. During training, with $\pi_{\theta}$ typically initialized from $\pi_{\text{ref}}$, the objective is to maximize the expected reward margin between $\mathbf{y}_w$ and $\mathbf{y}_l$:
\begin{equation} \label{eqn:dpo_loss}
    L_{\text{DPO}} = -\mathbb{\mathop{E}}_{(\mathbf{x},\mathbf{y}_w,\mathbf{y}_l)\sim D}[ \text{log}\sigma ( M(\mathbf{y}_w, \mathbf{y}_l, {\bf x}, \theta)  ) ]
\end{equation}
where the reward margin  $M(\mathbf{y}_w, \mathbf{y}_l, {\bf x}, \theta)$ is
\begin{equation}
\beta \, (\text{log}\frac{\pi_{\theta}(\mathbf{y}_w|\mathbf{x})}{\pi_{\text{ref}}(\mathbf{y}_w|\mathbf{x})} 
 -  \text{log}\frac{\pi_{\theta}(\mathbf{y}_l|\mathbf{x})}{\pi_{\text{ref}}(\mathbf{y}_l|\mathbf{x})} )
 \label{eq:dpomargin}
\end{equation}


\subsection{Related Work in Translation}
Previous work has explored the effectiveness of enhancing the translation performance of LLMs via Reinforcement Learning (RL) algorithms or supervised fine-tuning. \citet{dong2023raft} proposed RAFT that iteratively generates samples and fine-tunes the model on the filtered samples ranked by a reward model. \citet{gulcehre2023reinforced} proposed ReST that uses similar method for translation task, where they apply several fine-tuning steps on a sampled dataset, each time higher ranked samples.


Similar to our pairwise preference learning, \citet{zeng2023tim} introduced a framework TIM to enhance the translation performance of LLMs by learning to compare good translations and bad translations via a preference learning loss.

Contemporaneous with this work, \citet{finkelstein2023mbr} proposed MBR fine-tuning, which fine-tunes an NMT model on the MBR decoding outputs generated either by the model itself or by an LLM. However, their MBR fine-tuning utilizes only the final translations of MBR decoding whereas our fine-tuning method uses sets of sampled translations ranked by MBR, thus enabling the model to learn the same ranking preferences as MBR.


\section{Methodology}
Our method combines MBR decoding and DPO fine-tuning \cite{yang-2023-thesis}. We use the MBR procedure to calculate a score (Equation~\ref{eq:riskscore}) for each of a set of translation hypothesis generated by the base model. We then fine-tune the base model using the DPO objective (Equations \ref{eqn:dpo_loss},\ref{eq:dpomargin}) where the winning and losing hypotheses provided to DPO are chosen based on their relative MBR scores.    If successful, the fine-tuned model will have learned to rank translations  consistently with MBR decoding under the base model. 
%
\subsection{Creation of the DPO Preference Sets }
Following \citet{eikema-aziz-2022-sampling},  we use sampling to generate the translation hypotheses that will be used in DPO.  For a source sentence $\mathbf{x}$ we use simple ancestral sampling with a temperature of 0.7 to create a set of translations $H(x)=\{\mathbf{y}\sim\pi_{base}(\mathbf{y}|\mathbf{x})\}$  of size $|H(x)|$.   We use this collection as both the MBR evidence and hypothesis spaces \cite{GOEL2000115}.   

The hypotheses in $H(x)$ are ordered by their MBR scores as $\mathbf{y}_1, \mathbf{y}_2, ..., \mathbf{y}_{|H|}$ with the BLEURT metric~\cite{sellam-etal-2020-bleurt2} as the utility function.   The ordering reflects the MBR preference,  i.e. $\mathbf{y}_1$ would be the most preferred MBR hypothesis.    

\paragraph{Preference Selection Strategies}
DPO requires a set of preference triplets $\mathcal{D} = \{({\bf x},  \mathbf{y}_w, \mathbf{y}_l)\}$ where $\mathbf{y}_w$ has better MBR score than $\mathbf{y}_l$ and both of the hypotheses are selected from the hypothesis set $H(x)$. 
There are numerous strategies for selecting the preference pairs $(\mathbf{y}_w, \mathbf{y}_l)$ from the hypothesis set. We experimented with four selection schemes:
\begin{enumerate}
    \item \textbf{BW} is a simple strategy that selects the {\bf best} and {\bf worst} translation hypotheses from the ranked sets. For each source sentence $\mathbf{x}$, we only have one preference triplet $(\mathbf{x}, \mathbf{y}_1, \mathbf{y}_{|H(x)|})$.
    \item \textbf{BMW} adds the {\bf middle} hypothesis $\mathbf y_m$ from the ranked lists with index $m=\lceil |H(x)| / 2 \rceil$.   
This gives two triplets per source sentence: $(\mathbf{x}, \mathbf{y}_1, \mathbf{y}_{m})$ and $(\mathbf{x}, \mathbf{y}_{m}, \mathbf{y}_{|H(x)|})$.
    \item \textbf{CP} selects {\bf consecutive pairs} from the ranked list, yielding $|H(x)|-1$ triplets per source sentence, as   $(\mathbf{x},\mathbf{y}_1, \mathbf{y}_2)$, $(\mathbf{x},\mathbf{y}_2, \mathbf{y}_3)$, $\ldots$
\item \textbf{CPS} introduces a {\bf stride} into the CP selection strategy so as to avoid requiring DPO to learn distinctions between translations that are similarly ranked.       For example, with a {stride} of 2 we select triplets  $(\mathbf{x},\mathbf{y}_1, \mathbf{y}_3)$, $(\mathbf{x},\mathbf{y}_3, \mathbf{y}_5)$, $\ldots$
%
\end{enumerate}


\subsection{DPO Fine-Tuning}
With a set of preference triplets $\mathcal D$ selected by one of the schemes above,  
DPO fine-tuning proceeds as described in Section~\ref{sec:dpodescrip} and by \citet{DBLP:conf/nips/RafailovSMMEF23}.   The 
base model serves as the reference model in Equation~\ref{eqn:dpo_loss}. The base model is also used to initialise 
$\pi_{\theta}$, which is the model being fine-tuned.
The only DPO hyper-parameter we tune is $\beta$, which regulates how the fine-tuned model departs from the reference model~\citet{DBLP:conf/nips/RafailovSMMEF23}.


\section{DPO MBR Fine-Tuning and MT}


\begin{figure}[t]
    \centering
    \includegraphics[width=0.5\textwidth]{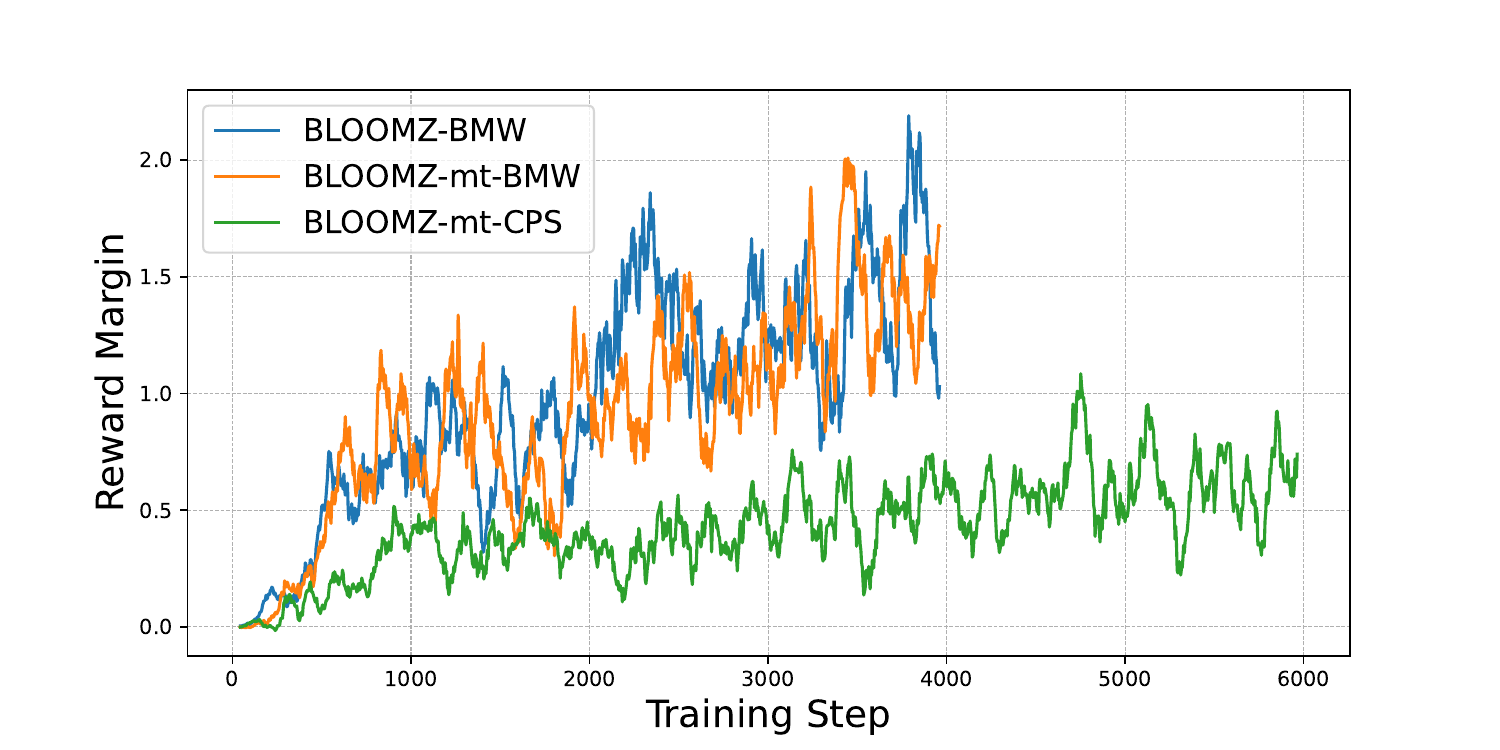}
    \caption{Reward margins for DPO MBR fine-tuning of BLOOMZ and BLOOMZ-mt with BMW and CPS (stride of 2) selection strategies. Margins are calculated on the Zh-En fine-tuning set (WMT20 test set) as fine-tuning  proceeds over one epoch. Results are plotted as moving averages with a window size of 20. CPS  yields more preference pairs than BMW.}
    \label{fig:margin_train}
\end{figure}

\noindent\textbf{Datasets}:   We evaluate translation on the WMT21 news translation test sets \cite{akhbardeh-etal-2021-findings} and the WMT22 general translation for Chinese-English \cite{kocmi-etal-2022-findings}, and the IWSLT 2017 test set for French-English \cite{cettolo-etal-2017-overview}. 
For DPO fine-tuning we use the source language text in the WMT20 test sets for Chinese-English \cite{barrault-etal-2020-findings} and IWSLT 2017 validation sets for French-English.    We do not use the corresponding reference translations, as  DPO MBR fine-tuning is unsupervised.   The fine-tuning and test sets are distinct and do not overlap.
%

\noindent\textbf{Models}: We use the BLOOMZ and BLOOMZ-mt models \cite{muennighoff-etal-2023-crosslingual} with 7.1 billion parameters as our base model. BLOOMZ-mt was pre-trained on 366 billion tokens from monolingual texts and was fine-tuned for translation task on Flores-200~\cite{nllbteam2022language} and Tatoeba~\cite{tiedemann-2020-tatoeba2} datasets. To prompt the model for translation, we include two randomly selected translation examples from the fine-tuning set into the input prompt as demonstration examples; these prompts are kept fixed throughout. In addition, we also fine-tuned the BLOOMZ-mt model in a supervised fashion for each language pair and denote this third base model as BLOOMZ-mt-sft. We use previous WMT news translation test sets from 2017 to 2020 as supervised fine-tuning sets for Chinese-English, and the first 20000 translation pairs from the IWSLT 2017 training set for French-English. Training details can be found in Appendix \ref{app: details}.

\noindent\textbf{Evaluation Metrics}: We use three evaluation metrics: BLEU \cite{papineni-etal-2002-bleu2}, BLEURT \cite{sellam-etal-2020-bleurt2}, and COMET-22 \cite{rei-etal-2020-comet, rei-etal-2022-comet}. 
BLEU serves only as a safety check:
Ideally DPO fine-tuning should not decrease BLEU. 


\noindent\textbf{Baselines and Targets}: We take the base model and evaluate it on all the test sets with both beam search and MBR decoding.  Our fine-tuned models, when decoded with beam search, should achieve similar performance as  MBR decoding under the base model and show improvement over the base model.
\begin{figure}[t]
    \centering
    \includegraphics[width=0.5\textwidth]{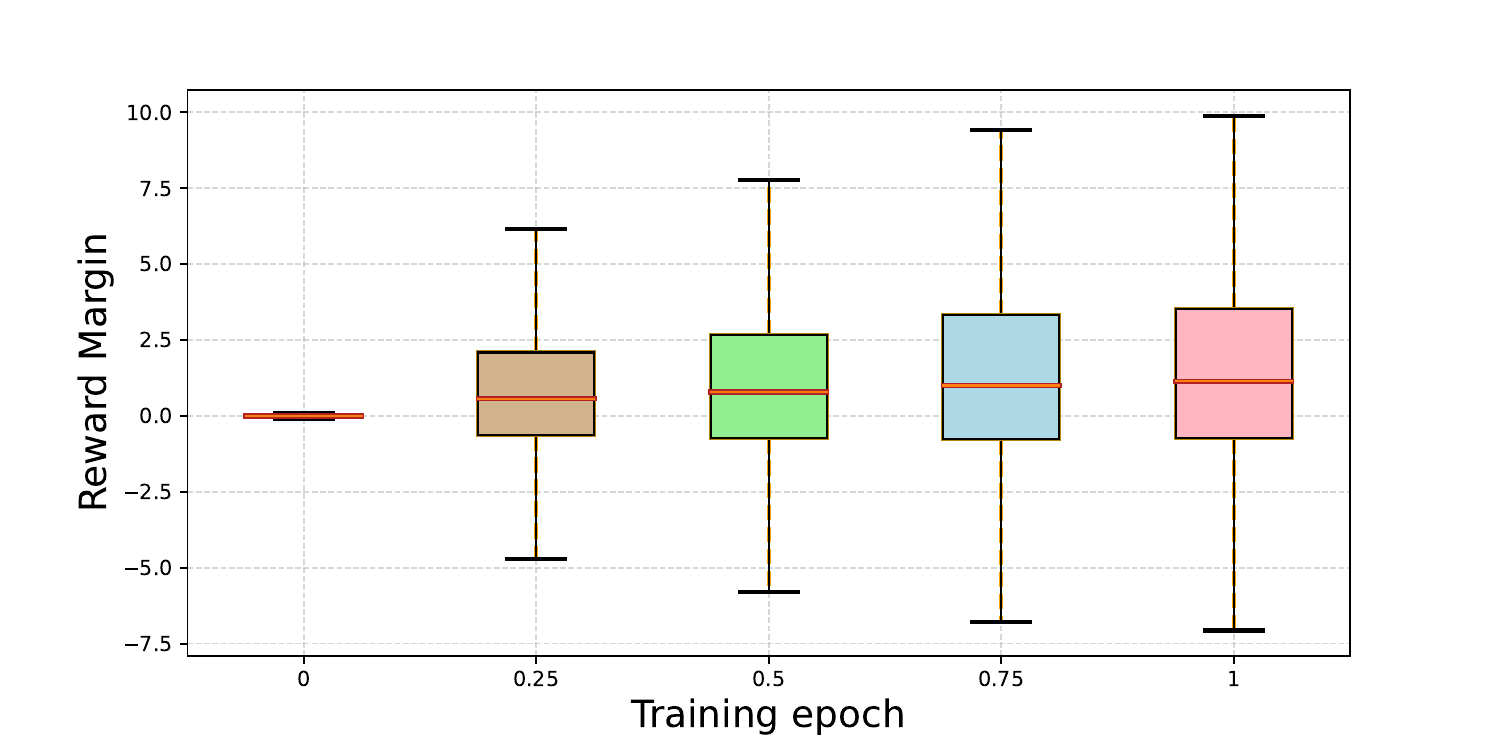}
    \caption{Reward margin distributions over all preference pairs extracted via the BMW scheme from a held-out dataset (WMT18 Zh-En test).  Distributions are gathered over the entire held-out  set at model checkpoints at the beginning, a quarter, middle, three quarters, and end of one epoch of DPO fine-tuning.  $|H| = 8$ and $\beta = 0.7$.  DPO fine-tuning generalises beyond its fine-tuning set and yields improved reward margins on held-out data.}
    \label{fig:margin_held}
\end{figure}
\begin{table*}[ht]
\small
    \centering
    \begin{tabular}{clcccccc}
    \toprule
    \multirow{2}{*}{\#} & \multirow{2}{*}{Model (Decoding)} & \multicolumn{2}{c}{WMT21} & \multicolumn{2}{c}{WMT22} & \multicolumn{2}{c}{IWSLT17}\\ \cmidrule(lr){3-4} \cmidrule(lr){5-6} \cmidrule(lr){7-8}
     & & zh-en & en-zh & zh-en & en-zh & fr-en & en-fr \\
     \midrule
     1 & BLOOMZ (Beam) & 59.6 | 76.5 & 59.2 | 81.1 & 59.9 | 74.6 & 55.9 | 76.7 & 72.7 | 83.9 & 69.3 | 83.1 \\
     2 & BLOOMZ (MBR $|H|=8$) & 60.0 | 76.4  & 62.5 | 82.3  & 62.1 | 75.8 & 62.7 | 80.0 & 73.6 | 84.2 & 70.4 | 83.3 \\ 
     3 & BLOOMZ (MBR $|H|=32$) & 62.5 | 77.2 & {\bf 64.7} | {\bf 83.0} &  {\bf 64.0} | 76.4 & {\bf64.9} | 80.7 & 74.8 | 85.0 & {\bf 72.6} | 84.3 \\
     4 & \textbf{BLOOMZ-DPO-MBR} (Beam) & {\bf 62.3} | {\bf 77.9} & 62.5 | 82.7 &{\bf 64.0} | {\bf 77.2} & 64.2 | {\bf 82.0} & {\bf76.5} | {\bf 86.9} & 72.2 | {\bf 84.8} \\
     \midrule
     5 & BLOOMZ-mt (Beam) & 60.3 | 77.0  & 59.2 | 80.9 & 60.9 | 75.5 & 59.0 | 79.1 & 74.8 | 85.4 & 70.3 | 83.5 \\
     6 & BLOOMZ-mt (MBR $|H|=8$) & 61.6 | 77.6 & 62.6 | 82.3 & 63.0 | 76.5 & 64.7 | 81.4 & 75.4 | 85.5 & 71.0 | 83.3 \\ 
     7 & BLOOMZ-mt (MBR $|H|=32$) & 63.4 | 78.3 & {\bf 64.9} | 82.9 & 64.8 | 77.2 & 66.8 | 82.1 & 76.3 | 86.0 & {\bf 73.2} | 84.3 \\
     8 & \textbf{BLOOMZ-mt-DPO-MBR} (Beam) & {\bf 63.9} | {\bf 78.7} & 64.0 | {\bf 83.6} & {\bf 65.1} | {\bf 77.9} & {\bf 67.6} | {\bf 83.7} & {\bf 76.5} | {\bf 86.8} & 71.9 | {\bf 84.6} \\
     \midrule
     9 & BLOOMZ-mt-sft (Beam) & 64.3 | 79.4 & 62.6 | 83.0 & 62.6 | 76.5 & 65.6 | 83.1 & 76.9 | 86.6 & 71.2 | {\bf 83.8} \\
     10 & BLOOMZ-mt-sft (MBR $|H|=8$) & 65.3 | 79.8 & 64.8 | 83.9 & 65.4 | 78.2 & 69.1 | 84.2 & 77.3 | 86.7 & 72.6 | 83.6 \\
     11 & BLOOMZ-mt-sft (MBR $|H|=32$) & {\bf 66.8} | 80.4 & {\bf 66.7} | {\bf 84.4} & {\bf 67.1} | 78.9 & {\bf 71.0} | 85.1 & {\bf 78.2} | {\bf 86.9} & {\bf 74.9} | 83.3 \\
     12 & \textbf{BLOOMZ-mt-sft-DPO-MBR} (Beam) & 66.0 | {\bf 80.8} & 64.2 | 83.9 & 66.5 | {\bf 79.6} & 69.5 | {\bf 85.6} & 76.4 | 83.4 & 72.4 | {\bf 83.8} \\
     \bottomrule
    \end{tabular}
    \caption{Translation performance in BLEURT and COMET (BLEURT | COMET) for models with beam search (beam width of 4) and MBR decoding on two language pairs from WMT21 news translation test sets, WMT22 general translation test sets, and IWSLT 2017 test sets. DPO-MBR indicates our translation performance with our fine-tuning method.   All the DPO MBR models were fine-tuned using the BMW strategy and $\beta=0.7$ except for BLOOMZ-mt-sft on IWSLT 2017, which used the BW strategy. We set $|H|=32$ to fine-tune BLOOMZ-mt-DPO-MBR on English-Chinese direction, $|H|=16$ on the French-English direction for BLOOMZ and BLOOMZ-mt, and set $|H|=8$ to fine-tune other DPO MBR models. DPO-MBR improves both BLEURT and COMET whenever MBR itself improves substantially over the baseline.   }
    \label{tab:main}
\end{table*}
We investigate two questions:   

(1) Can DPO teach MLLMs to learn their MBR translation preferences? 

(2) Does preference learning with DPO lead to improved translation?



\subsection{DPO Fine-Tuning Teaches a MLLM to Learn Its MBR Preferences}

Figure \ref{fig:margin_train} shows that the reward margins remain positive and, with some fluctuations, increase as fine-tuning proceeds, for all three models. This suggests that  DPO MBR fine-tuned models learn to put more probability mass on the winning hypotheses. The larger the margins, the more the models  prefer the winning over the  losing hypotheses.

To further investigate DPO MBR fine-tuning, we plot the distribution of reward margins on a held-out set, shown in Figure \ref{fig:margin_held}. The median of the distributions increase consistently as fine-tuning proceeds, indicating that the MBR preferences learned in fine-tuning also generalize to unseen data.

\subsection{DPO MBR Translation}
Table \ref{tab:main} gives our main translation results. Comparing Rows 3 \& 4, 7 \& 8, and 11 \& 12, we can see that DPO MBR fine-tuned models, when decoded with beam search, achieve similar performance in BLEURT and COMET as the base model decoded with MBR. The first two configurations (BLOOMZ-DPO-MBR and BLOOMZ-mt-DPO-MBR) outperform the base model's beam search results by $\approx 4$ BLEURT and $\approx 2$ COMET scores, and the third configuration outperforms the base mode by $\approx 3$ BLEURT and $\approx 2$ COMET on four out of six test sets. DPO MBR  improves the translation ability of BLOOMZ, BLOOMZ-mt across a range of test sets.  BLOOMZ-mt shows a  notable improvement after  DPO MBR fine-tuning, achieving the best performance in BLEURT on four out of six test sets and the best performance in COMET on all six test sets. We note that MBR decoding does not yield consistent improvement on the BLOOMZ-mt-sft model for IWSLT2017, and therefore does not provide a strong signal for DPO fine-tuning. We provide translation performance in BLEU in Appendix \ref{app: bleu} for reference.

\begin{table}[t]
\small
    \centering
    \begin{tabular}{ccccc}
        \toprule
        \# & $\beta$ & BLEU & BLEURT & COMET \\
        \midrule
        1 & (Baseline) & 16.4 & 60.3 & 77.0 \\
        \midrule
        2 & 0.1 & 9.9 & 64.5 & 71.3 \\
        3 & 0.3 & 11.8 & 64.8 & 73.5 \\
        4 & 0.5 & 14.3 & 64.0 & 76.1 \\
        5 & 0.7 & 16.4 & 63.3 & 77.7 \\
        6 & 0.9 & 17.6 & 61.8 & 77.9 \\
        \bottomrule
    \end{tabular}
    \caption{Effect of regularization parameter $\beta$ for DPO MBR fine-tuning of BLOOMZ-mt using  CPS with $|H|=8$. Models are fine-tuned on WMT20 zh-en and evaluated on WMT21 zh-en.}
    \label{tab:beta}
\end{table}

\subsubsection{KL-Divergence Regularization} 
We investigated the role of $\beta$,  the KL-divergence regularization factor, in DPO.  Table \ref{tab:beta} shows that fine-tuning with small $\beta$ values yields high BLEURT score (exceeding 64), but also a degradation in BLEU  and COMET.    Anecdotally, we find that small values of $\beta$ lead to repetitive outputs that are penalised heavily under BLEU and COMET.   
Gains in BLEU, BLEURT, and COMET are readily found, but 
we conclude that DPO MBR fine-tuning requires some care in regularization.

\begin{table}[t]
\small
    \centering
    \begin{tabular}{lccc}
    \toprule
        Selection Strategy & |H|=8 & |H|=16 & |H|=32 \\
        \midrule
        BW & 63.3 & 63.9 & 63.9 \\
        BMW & 63.9 & 64.2 & 63.6 \\
        CP & 62.5 & 62.4 & 60.4 \\
        CPS (strides of 2, 4, and 8) & 62.3 & 63.5 & 62.9 \\
        \bottomrule
    \end{tabular}
    \caption{WMT21 Zh-En BLEURT scores for  BLOOMZ with DPO MBR fine-tuning  with different preference pair selection strategies and hypothesis set sizes. The CP strategy results in lower performance in BLEURT compared to other strategies.}
    \label{tab:|H|}
\end{table}

\subsubsection{Effects of Pair Selection Strategy}
Table \ref{tab:|H|} shows that models trained on preference datasets constructed with the BW, BMW, and CPS pair selection strategies achieve similar performance on WMT21 Zh-En, with BLEURT scores in the range 62.9-63.9. DPO MBR appears robust to the selection of preference pairs. In terms of training efficiency, the BW and BMW strategies require fewer preference pairs (1 and 2 per source sentence, resp.) compared to the CPS strategy. However, these results show that some selection strategy is necessary since simply including all the pairs as in the CP strategy leads to degradation.

\subsubsection{Effects of Size of Hypothesis Set}
Table \ref{tab:|H|} shows that the number of hypotheses needed in the training preference dataset is less than that needed for MBR decoding (Rows 3 \& 7 in Table \ref{tab:main}). The best performance (BLEURT of 63.9) can be achieved with 16 hypotheses for the BW strategy and 8 hypotheses for the BMW strategy, an improvement over MBR decoding of the base model with $|H|=8$ (Row 2 \& 6 in Table \ref{tab:main}).




\section{Conclusion}
We introduce DPO MBR fine-tuning, an unsupervised preference optimization algorithm that leverages the ranked lists from MBR decoding to teach MLLMs the preference of MBR decoding. Our method enables MLLMs to achieve significant performance improvement when decoded with beam search in one pass, on par with the performance gained from two-pass MBR decoding\footnote{Codes are available at \url{https://github.com/BruceYg/DPO-MBR}}.

\section{Acknowledgement}
Jinghong Chen is supported by the Warwick Postgraduate Studentship from Christ's College and the Huawei Hisilicon Studentship for the undertaking of the PhD in Engineering at the University of Cambridge.

Weizhe Lin was supported by a Research Studentship funded by Toyota Motor Europe (RG92562(24020)). 

Prof. Bill Byrne holds concurrent appointments as a Professor of Information Engineering at Cambridge University and as an Amazon Scholar.  This publication describes work performed at Cambridge University and is not associated with Amazon.

We would also like to thank all the reviewers for their knowledgeable reviews.

\section{Limitations}

Our method was evaluated on WMT 2021 and WMT 2022 and IWSLT 2017 test sets, with high-resource languages only (English, Chinese, and French).  While our fine-tuned models performed well on these diverse test sets, behaviour may be different on medium-resource or low-resource languages or on other domains.

Our experiments focus on BLOOMZ and BLOOMZ-mt due to the ease of working with them and because BLOOMZ-mt is fine-tuned for translation.   Other (M)LLMs may yield different results.

We report MBR results using simple ancestral sampling.   Other work \cite{freitag2023epsilon} has found that there may be advantages in using other sampling schemes, such as epsilon sampling, for MBR.   Those other sampling methods potentially offer  further gains beyond what we have already shown.

 We do not report human assessments of translation quality to verify improvements, but we note that \citet{freitag-etal-2022-high} have reported extensive results showing that MBR decoding under BLEURT leads to improvements in translation quality as assessed by human judges.  We therefore take improvement in BLEURT as our main measurement of improved translation quality.  

\section{Risks}
Our unsupervised fine-tuning technique could potentially amplify undesirable biases or language already present in the baseline systems.  This could possibly happen if the MBR utility function,  in our case BLEURT, somehow encourages consensus amongst similar translations that are also undesirable.   Mitigation should be straightforward, in that any monitoring of the baseline models could also be applied after DPO MBR fine-tuning to reject fine-tuned models that exhibit any increase in bad behaviour.   Although it is not a focus of this work,  DPO MBR could possibly be used as a strategy for risk mitigation by penalizing undesirable behaviour through introduction of specific penalties into the MBR utility function.

\bibliography{anthology,custom}


\begin{table*}[!htbp]
\small
    \centering
    \begin{tabular}{clcccccc}
    \toprule
    \multirow{2}{*}{\#} & \multirow{2}{*}{Model (Decoding)} & \multicolumn{2}{c}{WMT21} & \multicolumn{2}{c}{WMT22} & \multicolumn{2}{c}{IWSLT17}\\ \cmidrule(lr){3-4} \cmidrule(lr){5-6} \cmidrule(lr){7-8}
     & & zh-en & en-zh & zh-en & en-zh & fr-en & en-fr \\
     \midrule
     1 & BLOOMZ (Beam) & 15.8 & 22.3 & 14.0 & 22.2 & 38.1 & 37.6 \\
     2 & BLOOMZ (MBR $|H|=8$) & 11.3  & 19.7  & 11.6 & 20.3 & 34.2 & 32.6 \\ 
     3 & BLOOMZ (MBR $|H|=32$) & 12.6 & 20.2 & 12.4 & 21.2 & 36.3 & 34.1 \\
     4 & \textbf{BLOOMZ-DPO-MBR} (Beam) & {\bf 17.2} & {\bf 23.7} & {\bf 15.6} & {\bf 26.5} & {\bf 40.6} & {\bf 38.9} \\
     \midrule
     5 & BLOOMZ-mt (Beam) & 16.4 & 22.5 & 14.7 & 26.2 & 38.7 & 37.8 \\
     6 & BLOOMZ-mt (MBR $|H|=8$) & 13.5 & 20.2 & 12.2 & 23.3 & 35.2 & 31.8 \\ 
     7 & BLOOMZ-mt (MBR $|H|=32$) & 14.3 & 20.8 & 13.0 & 24.0 & 36.9 & 33.8 \\
     8 & \textbf{BLOOMZ-mt-DPO-MBR} (Beam) & {\bf 18.0} & {\bf 22.7} & {\bf 15.9} & {\bf 26.9} & {\bf 40.4} & {\bf 38.3} \\
     \midrule
     9 & BLOOMZ-mt-sft (Beam) & 23.5 & {\bf 27.5} & 19.7 & 34.9 & {\bf 44.2} & {\bf 40.7} \\
     10 & BLOOMZ-mt-sft (MBR $|H|=8$) & 20.2 & 24.0 & 17.7 & 30.1 & 40.7 & 34.2 \\
     11 & BLOOMZ-mt-sft (MBR $|H|=32$) & 21.1 & 25.0 & 18.4 & 31.2 & 41.3 & 32.1 \\
     12 & \textbf{BLOOMZ-mt-sft-DPO-MBR} (Beam) & {\bf 23.8} & 26.3 & {\bf 22.1} & {\bf 35.4} & 27.3 & 38.5 \\
     \bottomrule
    \end{tabular}
    \caption{Translation performance in BLEU for models with beam search and MBR decoding on two language pairs from WMT21 news translation test sets, WMT22 general translation test sets, and IWSLT 2017 test sets. DPO-MBR indicates our translation performance with our fine-tuning method.   All the DPO MBR models were fine-tuned using the BMW strategy and $\beta=0.7$ except for BLOOMZ-mt-sft on IWSLT 2017, which used the BW strategy. We set $|H|=32$ to fine-tune BLOOMZ-mt-DPO-MBR on English-Chinese direction, $|H|=16$ on the French-English direction for BLOOMZ and BLOOMZ-mt, and set $|H|=8$ to fine-tune other DPO MBR models.}
    \label{tab:main_bleu}
\end{table*}

\newpage


\appendix

\section{Translation Performance in BLEU} \label{app: bleu}

In Table \ref{tab:main_bleu}, we provide the translation performance measured in BLEU score. The BLOOMZ-DPO-MBR and BLOOMZ-mt-DPO-MBR models achieve the best BLEU scores on all six test sets. The BLOOMZ-mt-sft model achieves lower BLEU score after DPO MBR fine-tuning on WMT21 English-to-Chinese, IWSLT17 French-to-English and English-to-French due to over-generation.

\section{Training Details} \label{app: details}

\subsection{DPO MBR Fine-tuning Details}
For DPO MBR fine-tuning, we trained each model for one epoch using the RMSProp optimizer. The learning rate is set to $5e^{-7}$ with 150 warmup steps. All fine-tuning experiments were done on two Nvidia A100-80G GPUs. We set the effective batch size to 4. We used FP32 and FP16 for the trained policy and the reference model in DPO fine-tuning, respectively.

\subsection{Supervised Fine-tuning}
We supervised fine-tuned the BLOOMZ-mt model on Chinese-to-English and English-to-Chinese directions for two epochs using previous WMT test sets. For French-to-English and English-to-French, we used the 20K translation pairs and trained for one epoch. Other hyper-parameters for SFT training are the same as DPO MBR fine-tuning.

\end{document}